\documentclass[12p]{article}

\usepackage[letterpaper,
            left=1.7in,
            right=1.7in,
            top=1.2in,
            bottom=1.2in]{geometry}

\usepackage{cite}
\usepackage{amsmath,amssymb,amsfonts}
\usepackage{graphicx}
\usepackage{textcomp}
\usepackage{mathabx}
\usepackage{hyperref}
\usepackage{xurl}

\usepackage[font=footnotesize,labelfont=bf]{caption}

\usepackage{amssymb}
\usepackage{amsmath}
\usepackage{amsfonts}
\usepackage{listings}
\usepackage{hyperref}
\usepackage{graphicx}
\usepackage{wrapfig}
\usepackage{xcolor}
\usepackage{mathtools}
\usepackage{algorithm}
\usepackage{algpseudocode}
\usepackage{textcomp}
\usepackage{cite}
\usepackage[caption=true]{subfig}

\newcommand{\C}{\mathcal{C}}

\newcommand{\cfreebar}{\mathcal{\Bar{\C}}_{\textup{free}}}

\newcommand{\qbar}{\Bar{q}}

\newcommand{\qstartbar}{\Bar{q}_{\textup{start}}}
\newcommand{\qgoalbar}{\Bar{q}_{\textup{goal}}}

\newcommand{\trrrt}{TR-RRT}

\newcommand{\R}{\mathbb{R}}

\title{Leveraging Two Robotic Arms \\ for Tight Assembly Performance Gains}

\author{Dror Livnat$^{\dagger}$ \and Yuval Lavi$^{\dagger}$ \and Michael M. Bilevich$^{\dagger}$ \and Dan Halperin$^{\dagger}$
}
\date{}

\begin{document}

\maketitle
\begingroup
\renewcommand\thefootnote{\fnsymbol{footnote}}
\footnotetext[2]{Blavatnik School of Computer Science and Artificial Intelligence, Tel-Aviv University, Israel. This work has been supported in part by the Israel Science
Foundation (grant no~3598/25), 
by the Blavatnik Computer Science Research Fund, and by the Shlomo Shmelzer Institute for Smart Transportation at Tel Aviv University.
}
\endgroup

\begin{abstract}
We provide a novel end-to-end framework for the execution of an assembly operation by two robotic arms, given the digital CAD models of the parts and their desired relative placement in their assembled state. We analyze and demonstrate the advantages of using two robotic arms simultaneously in tight assembly operations, compared to single-arm systems. 
Our method is implemented in both simulation and using physical robots. It provides theoretical guarantees on execution time and trajectory accuracy, supported by empirical evidence. 
In particular, we show that coordinated movement of two arms reduces average execution time by more than $50\%$ compared to using a single arm only, produces higher-quality trajectories, and accelerates the search for valid robot placements. Furthermore, we establish bounds on the required dimensions of the robotic cell.
Our open source software together with real-life video demonstrations are available in our project page. 
\end{abstract}

\begin{figure}[h]
\centering
\includegraphics[width=0.85\linewidth]{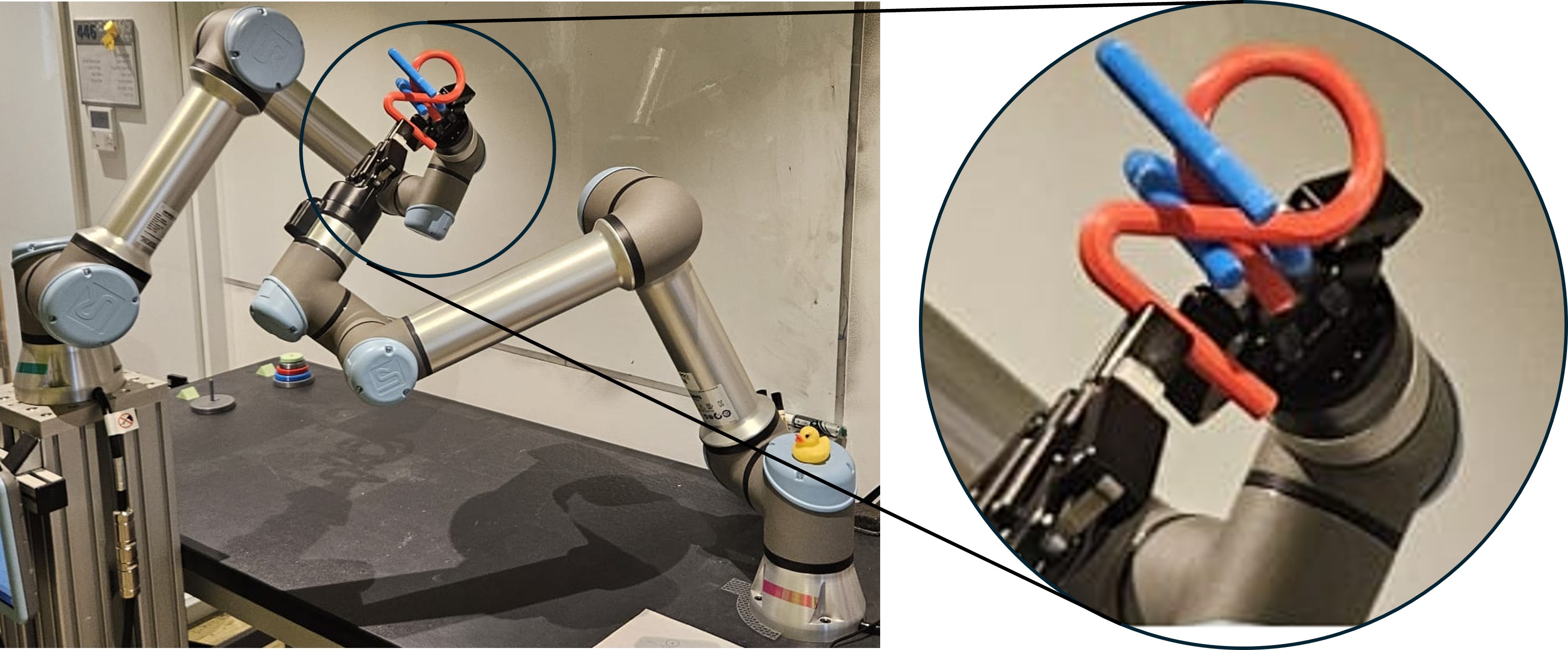}
\caption{Coordinated assembly using two robotic arms.}
\label{fig:az-zoom-in-out}
\end{figure}

\section{Introduction}
\label{sec:introcution}

In modern manufacturing, assembly consumes about $50\%$ of the total production time~\cite{ghandi2015assembly_planning_review, rashid2012review, fan_dong2003_20_50, YING2021assembly_arms}. Therefore, approaches that can significantly shorten assembly time have a strong impact on the industry. 
In this work, we present such an approach: a framework that translates CAD models into robotic arm motion plans, substantially accelerating assembly processes (Fig.~\ref{fig:az-zoom-in-out}).
We note that the present study focuses on one complete industrial assembly operation as a representative case, rather than an entire full-product assembly sequence, in order to isolate and address the fundamental planning challenges.

The growing demand for mass customization requires flexibility in production, often achieved during product assembly~\cite{michniewicz2016cad}. Consequently, assembly planning and robotic assembly have become prominent fields of research~\cite{ghandi2015assembly_planning_review, JIANG2022review_robotic_assembly}.

Industrial robots---especially in assembly---are widely acknowledged as central to Industry 4.0, driving improvements in flexibility, throughput, and quality~\cite{soori2024intelligent}. However, their widespread deployment is still hindered by the need for expert programming and complex system setup, which remains a significant bottleneck to agility and scalability~\cite{lohi2024programming, DBLP:journals/corr/abs-2106-07270}.

In a recent work~\cite{livnat2025fullcycle} we presented a framework for converting a digital design of assembly parts into a single robotic arm trajectory plan, together 
with the desired trajectory placement.

That work addressed three main challenges. First was the well studied free-flying objects motion planning~\cite{
kavraki1996probabilistic,
lavalle1998rapidly,
hsu2003bridge,
sun2005narrow,
karaman2011anytime,
klemm2015rrt,
salzman2016asymptotically,
thomas2018learning-tamar-abiel,
DBLP:journals/ral/KleinbortSLBH19,
DBLP:journals/tase/SalzmanHH15,
bency2019neural,
wang2021survey,
adamkiewicz2022vision,
tian2022assemble,
livnat2024tight}, which to date is still challenging in tight environments. Second was coping with the Continuous Path-Wise Inverse Kinematics (\emph{CPW-IK}), which converted a continuous assembly trajectory in the workspace into a continuous trajectory in the joint-space, overcoming challenges, such as singularities, different IK branches, and collision avoidance. The IK itself is also a well studied challenge~\cite{kucuk2014inverse, zhao1994inverse, robotics11060137, villalobos2022alternative, Schreiber21, ho2012efficient, LI2021104180, farzanehkaloorazi2018singularities, rakita2019stampede, wang2024iklinkendeffectortrajectorytracking}. However, no existing solution could handle the continuous task for UR-like arms efficiently enough to support trajectory-placement search. Hence, a dynamic programming based solution coupled with analytic IK~\cite{robotics11060137} was presented. 
Third was the challenge of \emph{trajectory placement}. This is dual to the well-studied \emph{robot placement} problem~\cite{makhal2018reuleaux}, in which feasible base locations are identified for each desired pose, or set of poses, of the robot end effector. Further work was done on mapping and testing robot placement for multiple isolated configurations~\cite{makhal2018reuleaux} in order for the robot to be able to complete a path. However, the challenge solved in~\cite{livnat2025fullcycle} was to place a full continuous trajectory in a complex $6D$ environment. 

Other works have provided a full cycle assembly planning for simpler tasks using a single robotic arm~\cite{wan2018assembly, tang2024automate, livnat2025fullcycle}, 
or construct motion plans, such that the transformation between its end effectors
remains fixed~\cite{10610675}. Using two arms can accelerate execution of tabletop rearrangement task as shown in~\cite{9357998}. 
However, we are not aware of any prior method that plans complex workspace assembly trajectories for two robotic arms.
 
In this work, we introduce such a framework, complemented by theoretical guarantees and experimental validation. Our results demonstrate clear benefits over single-arm systems: substantially shorter execution times, increased flexibility, and improved accuracy.

\noindent\textbf{This paper makes the following contributions:}
\begin{itemize}
    \item We introduce a new framework and two algorithms, \textbf{CPW-IK} and \textbf{GPW-IK}, that generate coordinated two-arm assembly plans directly from CAD models.
    \item We present a first theoretical analysis proving that average assembly operation time, using the new GPW-IK algorithm, is reduced by over $50\%$.
    \item We provide extensive experimental validation demonstrating significant gains in assembly time, trajectory accuracy, and algorithm runtime relative to single-arm baselines.
    \item We validate the approach using real robotic hardware, demonstrating practical feasibility.
\end{itemize}

Our open source software and real-life demonstrations are available on our project website:  \url{https://www.cgl.cs.tau.ac.il/projects/advantages-of-using-two-robotic-arms-for-tight-assemblies/}.

\section{Preliminaries and Problem Statement}
\label{sec:preliminaries}

In this section we summarize tools from~\cite{livnat2025fullcycle} needed for tight motion planning of free-flying objects, as well as motion planning using a single robotic arm. We then state the problem that we study in this work.
\subsection{Notation}
We briefly summarize the notation used throughout the paper. 
Configurations in the \emph{workspace} (i.e., rigid-body poses in $\mathrm{SE}(3)$) 
are denoted with a bar, e.g., $\bar{q} \in \mathrm{SE}(3)$. 
Joint configurations in the \emph{configuration space} of a robotic arm are written without a bar, e.g., $c \in \mathbb{R}^6$ for a $6$-DoF manipulator. 
We denote by $C_{\text{free}} \subseteq \mathbb{R}^6$ the collision-free subset of the joint configuration space, and $\Bar{C}_{free} \in \mathrm{SE}(3)$ its workspace equivalent. 
Paths are written as continuous functions $\gamma:[0,1] \to C_{\text{free}}$ in joint space and $\bar{\gamma}:[0,1] \to \mathrm{SE}(3)$ in workspace. 
When considering two arms, we use $C^2 = C^1 \times C^2$ to denote the combined configuration space.

\subsection{Tight Motion Planning of Free-Flying Objects}
\label{ssec:free-flying}
The input to this phase is a free-flying assembly problem. Specifically, we are given digital models of two free-flying bodies, $B_1$ and $B_2$, together with start and goal configurations $\qstartbar$ and $\qgoalbar$ for one of the bodies. The other body is assumed to remain fixed at the origin.
We deploy an appropriate motion planning algorithm, e.g., as described in~\cite{zhang2020cspace, tian2022assemble, livnat2024tight}. Specifically, in this work we use the~\trrrt\ algorithm from~\cite{livnat2024tight} as it best performs on the dataset presented in~\cite{tian2022assemble}, which is also in use in the current work. 

Most sampling-based algorithms, including those in~\cite{zhang2020cspace, tian2022assemble, livnat2024tight} suffer from two undesired properties. 
First, they usually allow for a small penetration between the sub-assemblies. This works well in simulation, but is challenging to realize in the physical world. Second, the output trajectories might be, as is often the case with sampling-based plans, unnecessarily long and jagged. We therefore follow the free-flying motion planning algorithm with a post processing phase. This includes the  \emph{retract} function from~\cite{livnat2024tight}, to remove the penetrations, followed by a smoothing function as described in~\cite{livnat2025fullcycle}. 

The output of this phase is a collision free, piecewise linear path $\Bar{\gamma}$,
from $\qbar_{\textrm{start}} \in \cfreebar$ to $\qbar_{\textrm{goal}} \in \cfreebar$, i.e., a path $\Bar{\gamma}:[0,1] \to \cfreebar$ such that $\Bar{\gamma}(0) = \qstartbar$ and $\Bar{\gamma}(1) = \qgoalbar$. The full sequence is

\vspace{-0.5 cm}
 \begin{equation}
     \qstartbar=\qbar_0,\ \qbar_1,\ \dots,\ \qbar_{N-1},\ \qbar_N = \qgoalbar,
 \end{equation}
and $\Bar{\gamma} \in \cfreebar$ consists of the segments between each pair of consecutive configurations in the sequence. 

\subsection{Continuous Path-wise IK for a Single Robotic Arm}
\label{ssec:pathwiseIK}

In this section, we summarize the method presented in~\cite{livnat2025fullcycle}. One major difficulty of transferring a trajectory from $\mathrm{SE}(3)$ to the joint-angle space is finding a consistent branch of inverse-kinematics~\cite{Schreiber21}. 
For each end-effector pose in $\mathrm{SE}(3)$, the six-DoF UR-like robotic arm has up to eight possible different inverse-kinematics solutions, which can be analytically computed. 

For each pose in the trajectory of a free-flying object a fixed pose of the respective end-effector is induced based on a predefined grasping.  
For that end-effector desired pose, all possible IK solutions are calculated. Then, 
a layered Directed Acyclic Graph (DAG) $G=(V,E)$ is constructed, where layer $L_k$ contains the feasible IK solutions of pose $k$ in the original trajectory. A node $a\in L_k$ is connected with a directed edge $e$ to node $b\in L_{k+1}$ if and only if their $L^1$ distance in the joint-angle space is smaller than some predefined threshold $\delta$. The weight of the edge is that same $L^1$ distance $w(e)=||a - b||_{L^1}$.
Two dummy nodes $s, \; t$ are added to the graph, and are connected with zero-weight edges to the nodes of layers $L_1$ and $L_N$ respectively, with $N$ being the number of poses in the original trajectory. Denote by $\Pi_G(s,t)$ all paths from $s$ to $t$ in the graph $G$.

The algorithm finds a path $\gamma$ whose heaviest edge has the smallest weight. That is, it chooses $\gamma$:
\begin{align}
    \gamma = {\arg\min}_{\rho\in\Pi_G(s,t)}\left\{ \max_{e\in\rho} \;  w(e) \right\}.
\end{align}

The \emph{error} of the trajectory $\gamma$ is defined as follows: Let $p\in\partial A$ be a point on the boundary of the sub-assembly manipulated by the robotic arm. Let $\gamma_p, \Bar{\gamma}_p: [0,1] \to \mathbb{R}^3$ be the trajectories of $p$ after following the paths $\gamma$ and $\Bar{\gamma}$, respectively. 
The one-sided Hausdorff distance between $\gamma_p$ and $\Bar{\gamma}_p$ is $d(\gamma_p, \Bar{\gamma}_p) = \max_{x \in \gamma_p} \min_{y\in \Bar{\gamma}_p} ||x-y||$. Then the error of the path $\gamma$ is the largest one-sided Hausdorff distance of any point $p\in\partial A$:
\begin{align}
    \mathrm{Err}(\gamma) = \max_{p\in \partial A}\;   d(\gamma_p,\ \Bar{\gamma}_p)\;.
\end{align}
Our goal is to bound this error by some arbitrarily small $\varepsilon$.

Denote by $D$ the maximal distance of a point in the moving system (the robot and the dynamic body attached to it) from the robot's base. Then, it is shown in~\cite{livnat2025fullcycle} that limiting $\delta_{\gamma}$, the maximal weight of an edge along the trajectory, $\delta_\gamma \leq \frac{\varepsilon}{D}$, yields the desired error bound $\mathrm{Err}(\gamma) \leq \varepsilon$.

\subsection{Trajectory Placement for a Single Robotic Arm}

For a single robotic arm, choosing merely the initial pose of the dynamic sub-assembly $B_2$, which is referred as \emph{placement}, determines the entire placement of the trajectory. Note that not all placements have a valid corresponding trajectory in joint-angle space.
Hence, for a single robotic arm, we begin by randomly sampling a six-dimensional starting pose, and run the CPW-IK algorithm. If successful, a valid placement for the trajectory is found; otherwise, we sample a new initial pose and repeat the process.

\subsection{Problem Statement}
\label{ssec:problem}

The input to the framework consists of CAD models of two rigid bodies to be assembled, together with desired start and goal relative poses in $\mathrm{SE}(3)$. The output is detailed instructions, in joint-angle space, of how each of two robotic arms should move in order to perform the assembly. This includes placement of the pieces at the beginning of the assembly, grasping poses, and a \emph{time-parameterized trajectory} of each robotic arm. Formally, we start with two rigid bodies $B_1, B_2 \subseteq \R^3$, which we wish to assemble. We assume that the bodies are placed in a pose aligned with their grasp by a robotic arm.
In the CAD software however, they may be posed differently, with offsets of $\Bar{q}_{B_1}, \Bar{q}_{B_2} \in \mathrm{SE}(3)$ for $B_1, B_2$, respectively. Hence, when planning for free-flying objects, we can assume that $B_1$ is fixed at the pose $\Bar{q}_{B_1}$, and that $B_2$ is a dynamic object that we wish to move from $\Bar{q}_{B_2}$ to $\qgoalbar \cdot \Bar{q}_{B_2}$.

The configuration space (C-space) of a single UR-like robotic arm is $\mathcal{C} = [0,4\pi)^6 \subseteq \R^6$. The forward kinematics function $f:\mathcal{C} \to \mathrm{SE}(3)$ maps a configuration $c \in C$ to its corresponding workspace pose $\qbar = f(c)$ of the end-effector. 

When two robotic arms $R_1, R_2$ are in play, the combined configuration space is the product $\mathcal{C}^2=\mathcal{C}\times \mathcal{C} \subset \mathbb{R}^{12}$. That is $c^2 = (c^{(1)},\ c^{(2)}) \in \mathcal{C}^2$ is a pose such that the arm $R_1$ is at configuration $c^{(1)}$ and arm $R_2$ at configuration $c^{(2)}$.

We say that $c^2 \in \mathcal{C}^2_{forbid}\subseteq \mathcal{C}^2$
if any two of $R_1, R_2, B_1, B_2$ are in intersection among themselves or with an obstacle.

We define the \emph{free region} as all configurations that yield no intersection of the interiors of the robots, the objects, and the environment obstacles:
\begin{align}
    \mathcal{C}^2_{free} = \left(\mathcal{C}^2 \setminus \mathcal{C}^2_{forbid}\right).
\end{align}

We are now ready to define our problem.

\paragraph*{Problem Statement} 
We are given CAD models of two rigid bodies $B_1, B_2 \subseteq \mathbb{R}^3$, their initial configurations $\Bar{q}_{B_1}, \Bar{q}_{B_2} \in \mathrm{SE}(3)$, and a desired goal pose $\qgoalbar$ for body $B_2$. 
The objective is to compute a collision-free motion path 
$\gamma = (\gamma^{(1)}, \gamma^{(2)}): [0,1] \to \mathcal{C}^2_{\textrm{free}}$ 
that satisfies the following conditions:
\begin{align}
   f\left(\gamma^{(1)}(0)\right)^{-1} \cdot 
        f\left(\gamma^{(2)}(0)\right) &= 
            (\Bar{q}_{B_1})^{-1} \cdot \Bar{q}_{B_2}\;, \label{eq:start_constraint}\\
    f\left(\gamma^{(1)}(1)\right)^{-1} \cdot 
        f\left(\gamma^{(2)}(1)\right) &= 
            (\Bar{q}_{B_1})^{-1} \cdot (\qgoalbar \cdot \Bar{q}_{B_2})\;.\label{eq:goal_constraint}
\end{align}
These constraints ensure that the relative configuration of $B_2$ in the frame of $B_1$ matches the given initial conditions at $t=0$ (equation~\ref{eq:start_constraint}) and the desired goal configuration at $t=1$ (equation~\ref{eq:goal_constraint}).

We note that a choice for such a path $\gamma$ also incorporates a choice for the placement of the rigid bodies in the workspace. That is, $\gamma$ may be any valid trajectory in joint-angle space that starts with assembled pieces and ends with dis-assembled pieces.

\smallskip

In this work we restrict ourselves to paths that are piecewise linear. A path
 \begin{align}     
    \gamma = \{c^2_{\textrm{start}} = c^2_0,\ c^2_1,\ \dots,\ c^2_{N-1},\ c^2_N = c^2_{\textrm{goal}} \},
 \end{align}
is valid, if the segment between each pair of consecutive configurations in the sequence is contained in $C^{2}_{free}$.

\smallskip

Within this general definition of $\gamma$, we explicitly implement three types of trajectories. First is a single arm trajectory as presented in~\cite{livnat2025fullcycle}, where arm $R_1$ keeps body $B_1$ static, and arm $R_2$ does all the work with $B_2$. The second and the third, which are defined in the next section, are two different algorithms providing \emph{coordinated two-arm plan}, where at each step both of the arms may move.

\section{Method}
In this section we show how the single arm CPW-IK method presented in section~\ref{ssec:pathwiseIK} can be extended to the case of two arms, using two different approaches. The input to this phase is the output of the free-flying objects motion plan, which is a path $\Bar{\gamma}:[0,1] \to \cfreebar$ such that $\Bar{\gamma}(0) = \qstartbar$ and $\Bar{\gamma}(1) = \qgoalbar$, as discussed in section~\ref{ssec:free-flying}. 

To develop algorithms using two robotic arms, we refine the definition of a \emph{segment} as follows.
For the two-arm execution phase, segment $i$ is anchored by the \emph{absolute} world poses of the two bodies at its start,
\[
\sigma_i \;=\; (\bar{p}_{1,i},\ \bar{p}_{2,i}) \in \mathrm{SE}(3)\times \mathrm{SE}(3),
\]
together with the next \emph{relative} target pose $\bar{q}_{i+1}\in \mathrm{SE}(3)$ induced by the free-flying path.
Our relative-pose convention is
\[
\bar{q} \;=\; \bar{p}_1^{-1}\bar{p}_2,
\]
such that the segment begins at $\bar{q}_i=\bar{p}_{1,i}^{-1}\bar{p}_{2,i}$ and ends at any pair of absolute poses
$(\bar{p}_{1,i+1},\bar{p}_{2,i+1})$ satisfying the constraint
\[
\bar{p}_{1,i+1}^{-1}\bar{p}_{2,i+1} \;=\; \bar{q}_{i+1}.
\]
This freedom of choosing the absolute end pose of $B_1$ (and hence of $B_2$) while meeting the required relative target is the key ingredient that allows the coordinated two-arm execution-time gains.
In both algorithms we go over the free-flying trajectory $\Bar{\gamma}$ step by step, and use the end pose from segment $i$ as the starting pose for segment $i{+}1$.
We aim to find a corresponding collision-free path $\gamma:[0,\; 1] \to \mathcal{C}^2$ in the two robotic arms joint space.

\smallskip
\noindent\textbf{Execution-time model.}
For each motion segment, we assume synchronized joint execution: the joint(s) with maximal angular displacement move at a fixed maximal velocity, and the remaining joints move at constant lower velocities so all joints complete the segment together. Hence, segment time is proportional to the maximum joint angular displacement in that segment.

\begin{figure}[t!]
\centering
\includegraphics[width=1.0\linewidth]{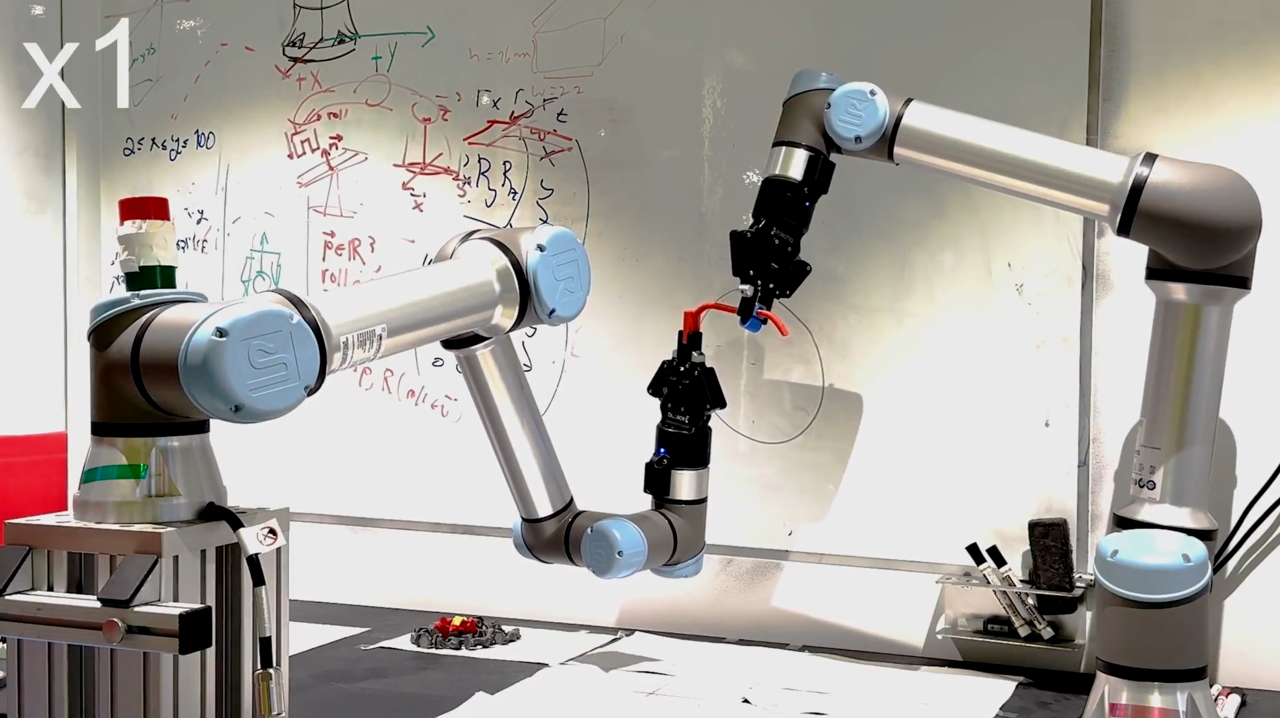}
\caption{Coordinated assembly of industrial assembly number 16505.}
\label{fig:16505}
\end{figure}

\subsection{CPW-IK for Coordinated Two-Arms}
\label{ssec:cpw-ik-dual-arm}

First, we sample a random placement for object $B_1$. That is, we sample a random $\Bar{q}_{rand} \in \mathrm{SE}(3)$. 

For each step $\Bar{q}_i \mapsto \Bar{q}_{i+1}$ in the desired trajectory $\Bar{\gamma}$ we do as follows:
\begin{itemize}
    \item Calculate the midpoint between $\Bar{q}_i$ and $\Bar{q}_{i+1}$.
    \item Find all IK solutions for that midpoint pose, and add those solutions as a layer in the dynamic DAG, for the object $B_2$.
    \item Find the corresponding pose for part $B_1$ in $\mathrm{SE}(3)$ by multiplying the new pose of $B_2$ by the transformation matrix from that same pose in $\Bar{\gamma}$ and $\Bar{q}_{B_1}$. Find all possible IK solutions for that pose and add the layer to the DAG, for the rigid part $B_1$.
\end{itemize}

Each edge in these layers, corresponds to simultaneous movement of the robotic arms $R_1$ and $R_2$ towards that midpoint. After building the DAG, we use the same dynamic programming presented in~Section~\ref{ssec:pathwiseIK} to select a path minimizing the one-sided Hausdorff distance. That is  $d_H(\{ f(\gamma^{(1)}(t))^{-1} f(\gamma^{(2)}(t)) : t \in [0, 1] \}, \{q^{-1}_{B_1} \Bar{\gamma}(t) : t \in [0, 1]\})$, the distance between the realized trajectory of $B_2$ in the frame of $B_1$ obtained via forward kinematics and the desired trajectory of $B_2$ specified in the input. Alternatively, we minimize the makespan of the path.

\subsection{Greedy Path-Wise IK for Coordinated Two-Arms}

We now present the Greedy Path-Wise Inverse Kinematics (GPW-IK) algorithm, our new coordinated planning algorithm, which is the first to guarantee more than $50\%$ reduction in per-average-segment execution
time compared to independent arm motions.
Throughout this section, $\bar{q}_i$ denotes the \emph{relative configuration of $B_2$
in the frame of $B_1$}, with the convention $\bar{q}_i=\bar{p}_{1,i}^{-1}\bar{p}_{2,i}$.
Thus, a segment is defined by the relative update $\bar{q}_i \mapsto \bar{q}_{i+1}$, while the absolute end poses
$(\bar{p}_{1,i+1},\bar{p}_{2,i+1})$ are free to vary subject to $\bar{p}_{1,i+1}^{-1}\bar{p}_{2,i+1}=\bar{q}_{i+1}$. Although segments are indexed by relative configurations
$\bar q_i \to \bar q_{i+1}$, a segment realization is defined by a transition
between absolute world poses $(\bar p_{1,i}, \bar p_{2,i}) \to
(\bar p_{1,i+1}, \bar p_{2,i+1})$ that satisfies
$\bar p_{1,i}^{-1}\bar p_{2,i}=\bar q_i$ and
$\bar p_{1,i+1}^{-1}\bar p_{2,i+1}=\bar q_{i+1}$.

The GPW-IK algorithm synchronizes the execution 
of the two arms so that they complete the work on each segment simultaneously. 
Instead of splitting evenly as in CPW-IK, GPW-IK computes a per-segment split 
that minimizes the common completion time.

\paragraph{Per-segment rule}
Let $t_1$ denote the time required for $R_1$ to execute the entire segment alone 
(keeping $R_2$ static), and let $t_2$ be the analogous time for $R_2$. 
GPW-IK chooses a split
\[
x = \frac{t_2}{t_1+t_2}, \qquad 1-x = \frac{t_1}{t_1+t_2},
\]
so that both arms complete their partial motions in the same synchronized time
\[
T = \frac{t_1 t_2}{t_1+t_2}.
\]

\paragraph{Toy example}
Suppose $t_1 = 6$\,s and $t_2 = 4$\,s. Then $x=0.4$ and $1-x=0.6$, yielding a 
synchronized execution time $T = 2.4$\,s. By comparison, CPW-IK’s even split 
would take $\max(0.5\cdot 6, \;0.5\cdot 4)=3$\,s, so GPW-IK achieves a shorter makespan.

\paragraph{Algorithm (per segment)}
For each segment $\bar{q}_i \mapsto \bar{q}_{i+1}$, GPW-IK performs:
\begin{enumerate}
  \item Take into account all possible (up to 8) IK solutions for each arm, and select the one that minimizes the time for that step for each arm, if it were to perform the segment alone. Compute the single-arm execution times $t_1$ and $t_2$ as above. 
  \item Compute the synchronization fractions 
        $x=\tfrac{t_2}{t_1+t_2}$ and $1-x=\tfrac{t_1}{t_1+t_2}$.
  \item \textbf{Partial move of $R_2$:} move each joint of $R_2$ by fraction 
        $1-x$ of the way toward the configuration that would realize 
        $\bar{q}_{i+1}$ if $R_1$ stayed fixed. 
  \item \textbf{Placement of $B_1$:} place $B_1$ so that $B_2$ is in the required relative configuration $\bar{q}_{i+1}$. More formally, if $p_i \in \mathrm{SE}(3)$ is the pose of $B_i$ in world coordinates, move $B_1$ to 
\[
\bar{p}_1^{\text{target}} \;=\; \bar{p}_2^{(\text{partial})} \, (\bar{q}_{i+1})^{\; -1}.
\]
  \item \textbf{Move of $R_1$:} calculate the IK for $\bar{p}_1^{\text{target}}$ and move $R_1$ accordingly.
  \item Proceed to the next segment.
\end{enumerate}

\paragraph{Comparison to CPW-IK}
CPW-IK enforces an equal (50/50) split, which is only optimal if $t_1=t_2$. 
GPW-IK adapts the split per segment: if $t_1\neq t_2$ it assigns more of the 
motion to the faster arm and strictly improves the makespan. 
If $t_1=t_2$, the two methods coincide.

\section{Analysis and Guarantees}
\subsection{GPW-IK Analysis}
We analyze the execution time of GPW-IK under the standard joint-velocity model used throughout this paper: for each segment, the joint requiring the largest angular displacement moves at its maximal constant speed, while all other joints move at equal or lower constant speeds. Consequently, the execution time of a segment is proportional to the maximum joint angle traversed along that segment. Under this model, we compare the synchronized two-arm execution of a segment to its single-arm counterpart.

In this section, $\bar{q}_i$ denotes the \emph{relative configuration of $B_2$ 
in the frame of $B_1$}. A segment is therefore defined by a change 
$\bar{q}_i \mapsto \bar{q}_{i+1}$.

In order to estimate the expected makespan improvement that GPW-IK provides over the usage of a single arm, we look at all possible short segments in the workspace, and compare the time it takes two robotic arms using GPW-IK to complete the motion along such a segment, to the time it would have taken a single arm.

Consider such a short segment 
$\textit{s}:=\Bar{q}_i \mapsto \Bar{q}_{i+1}$, and consider two identical robotic arms in a shared workspace. For a given motion segment, or step:
\begin{itemize}
\item Arm $R_1$ requires $t_1 > 0$ time to execute the entire motion.
\item Arm $R_2$ requires $t_2 > 0$ time to execute the entire motion.
\end{itemize}
In GPW-IK we let the arms execute complementary fractions of the motion such that they finish simultaneously. 

Let $x$ be a fraction of the motion performed by arm $R_1$. That is, if $(r_1, r_2, ..., r_6) = c^{(1)}_{i+1} - c^{(1)}_i$ is the set of rotations required by $R_1$ in order to move the assembly from $\Bar{q}_i$ to $\Bar{q}_{i+1}$, then moving a fraction $x$ is making the step $x \cdot (r_1, r_2, ..., r_6)$.
Then the times of arms $R_1$ and $R_2$ are $x t_1$ and $(1 - x) t_2$ respectively.
For simultaneous completion:
\[
x t_1 = (1 - x) t_2,
\]
which gives:
\[
x = \frac{t_2}{t_1 + t_2} \text{ and } 1 - x = \frac{t_1}{t_1 + t_2}.
\]
The common completion time is:
\[
T_\textit{two-arms} = x t_1 = (1 - x) t_2 = \frac{t_1 t_2}{t_1 + t_2}
\]

By symmetry over the workspace, there exists an equivalent segment where the times are swapped ($t_1 \longleftrightarrow t_2$). For that segment, the optimal completion time is the same:
\[
T_\textit{two-arms} = \frac{t_1 t_2}{t_1 + t_2}.
\]
Thus, the time for completion of both segments combined in a two-arm setting:
\[
T_\textit{total, two-arms} = 2 \cdot \frac{t_1 t_2}{t_1 + t_2}.
\]

In a single arm setting, the total time it takes a single arm $R_1$ to execute both segments alone is:
\[
T_\textit{total, single} = t_1 + t_2.
\]
The time ratio (two-arms setting vs. single-arm setting) for completion of both segments is then:
\[
\rho = \frac{T_\textit{total, two-arms}}{T_\textit{total, single}} = \frac{2 \cdot \frac{t_1 t_2}{t_1 + t_2}}{t_1 + t_2} = \frac{2 t_1 t_2}{(t_1 + t_2)^2}.
\]

When $t_1 = t_2$, we have:
\[
\rho = \frac{2 t_1^2}{(2 t_1) ^2} = \frac{1}{2}.
\]
This is the maximum possible ratio. For unequal times $t_1 \ne t_2 > 0$ we have from the Arithmetic Mean-Geometric Mean inequality:
\[
\frac{t_1 + t_2}{2} \ge \sqrt{t_1 t_2}.
\]

Squaring both sides and rearranging yields an upper bound for $\rho$:
\[
\rho = \frac{2 t_1 t_2}{(t_1 + t_2)^2} \le \frac{1}{2}.
\]

This analysis shows that, on average, the execution time for a short segment is reduced by more than $50\%$ compared to single-arm execution. This is also true for \emph{average} collections of such segments. Although this gives hope to find trajectories for two robotic arms that save more than $50\%$ of the time, we do not claim that it guarantees that the \emph{optimal} two-arm solution is shorter by more than $50\%$ than the \emph{optimal} single arm solution, as they do not consist of the same collection of segments. We therefore ran a large series of experiments to assess the practical strength of the algorithm.

\subsection{Path Accuracy}

Here we briefly outline how the guarantee for path accuracy developed in~\cite{livnat2025fullcycle} and reviewed in \ref{ssec:pathwiseIK} can be extended to the two-arm case. Let $p\in\partial A$ be any point on the boundary of an assembly part. 
Consider a segment 
$(c^2_i,\ c^2_{i+1})$. Its endpoints, $c^2_i$
and $c^2_{i+1}$ serve as anchor points where, by construction, the error, or the distance between any point $p$ and where it should have been according to the free-flying trajectory $\Bar{\gamma}$ is zero (see step $4$ in the GPW-IK algorithm). 

We begin by redefining $D$: instead of denoting the distance from the base of an arm to the farthest point on the single assembly piece it manipulates, $D$ now represents the distance from the base of an arm to the farthest point on any of the assembly pieces. Recall that the resolution of the input free-flying trajectory $\Bar{\gamma}$ is such that the maximal angle required to traverse a segment $q_i \mapsto q_{i+i}$ is some $\delta$. In GPW-IK, the sum of the maximal angles $\alpha_1,\ \alpha_2$ traversed by $R_1, R_2$ respectively along a single segment  is therefore smaller than that $\delta$, $\alpha_1 + \alpha_2 < \delta$. 
For the first half of the motion within a segment, we therefore have $\alpha_1 + \alpha_2 \le \frac{\delta}{2}$. In this setting, the maximum displacement of a point $p \in\partial A$ in $\mathbb{R}^3$ is bounded by $\frac{\alpha_2}{2} \cdot D$. 
At the same time, its reference point $p'$, initially coincident with $p$ and displaced by arm $R_1$, moves by at most $\frac{\alpha_1}{2} \cdot D$. Hence, the maximal discrepancy between $p$ and $p'$ is bounded by $(\frac{\alpha_1}{2} + \frac{\alpha_2}{2}) \cdot D \le \frac{\delta}{2} \cdot D$. 
By symmetry, and since $c^2_{i+1}$ is also an anchor point where the distance between any point $p$ and its reference point $p'$ is zero, the same reasoning applies throughout the entire segment. Consequently, imposing the condition $\delta \le \frac{\epsilon}{D}$, as in the single-arm case, ensures that the maximal one-sided Hausdorff distance of any point $p$ is $\mathrm{ERR}(\delta) \le \frac{\epsilon}{2}$, i.e., exactly half the error guarantee of the single-arm scenario in~\cite{livnat2025fullcycle}.

\begin{figure*}[h] 
  \centering \includegraphics[width=\textwidth ]{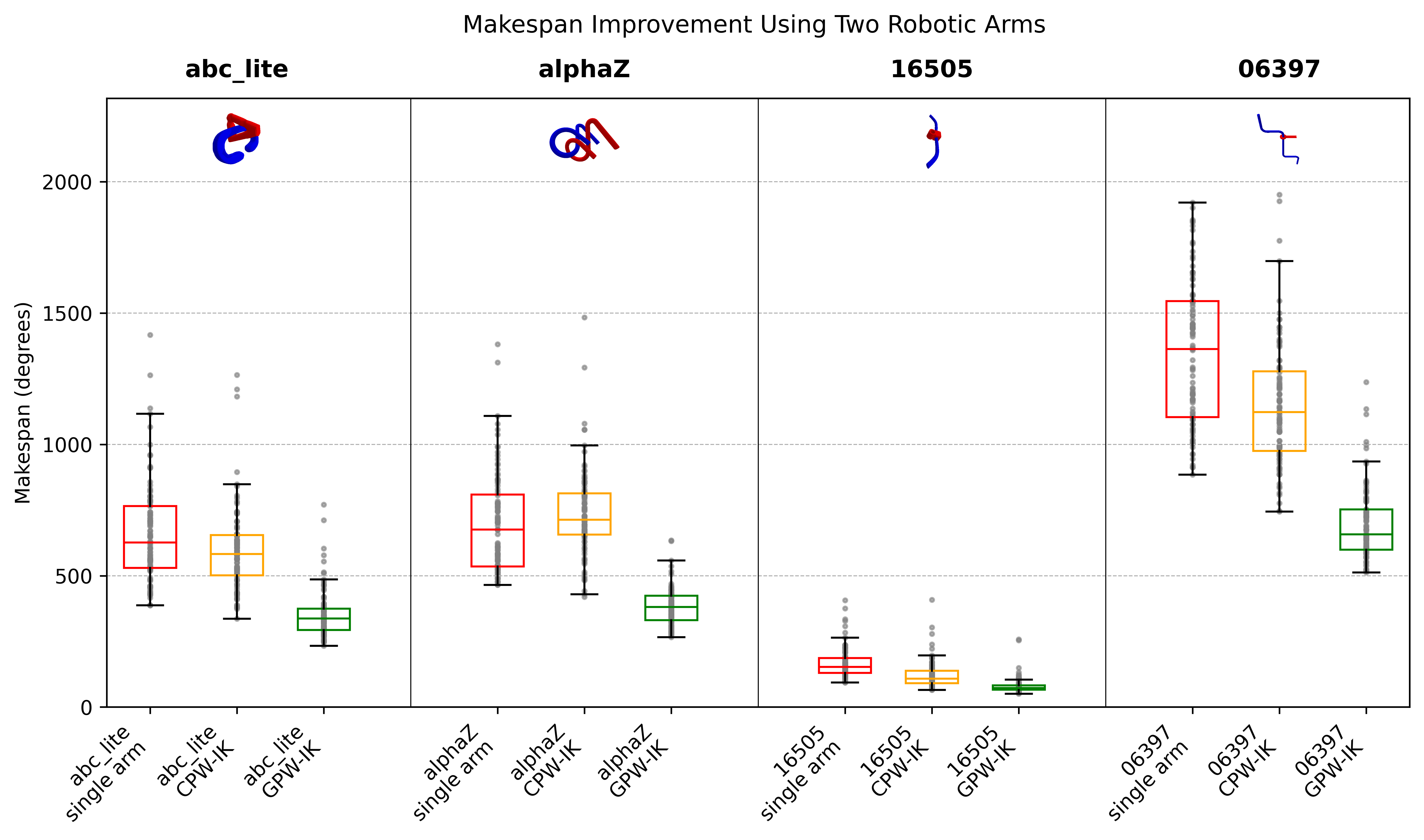} 
  \caption{Makespan (total execution time) for each assembly and each IK algorithm, evaluated over 100 valid placements per algorithm–assembly pair.
The metric is the sum, across all trajectory segments, of the largest joint rotation (degrees) in each segment.
Because execution time scales linearly with joint angular speed, this measure is proportional to actual runtime.}
  \label{fig:makespan}
\end{figure*}

\subsection{Robotic Cell Size Guarantee}

In this section, we provide guarantees regarding the space that the two-arm robotic cell will occupy. This is based solely on the components involved and the positions of the robotic arms, regardless of the specific trajectory and placement. 

Understanding the dimensions of the robotic cell is crucial when deploying robotic arms in factories and workshops. It ensures safety, facilitates collision avoidance, and optimizes the use of floor space.

\begin{figure}[H]
\centering
\includegraphics[width=0.8\linewidth]{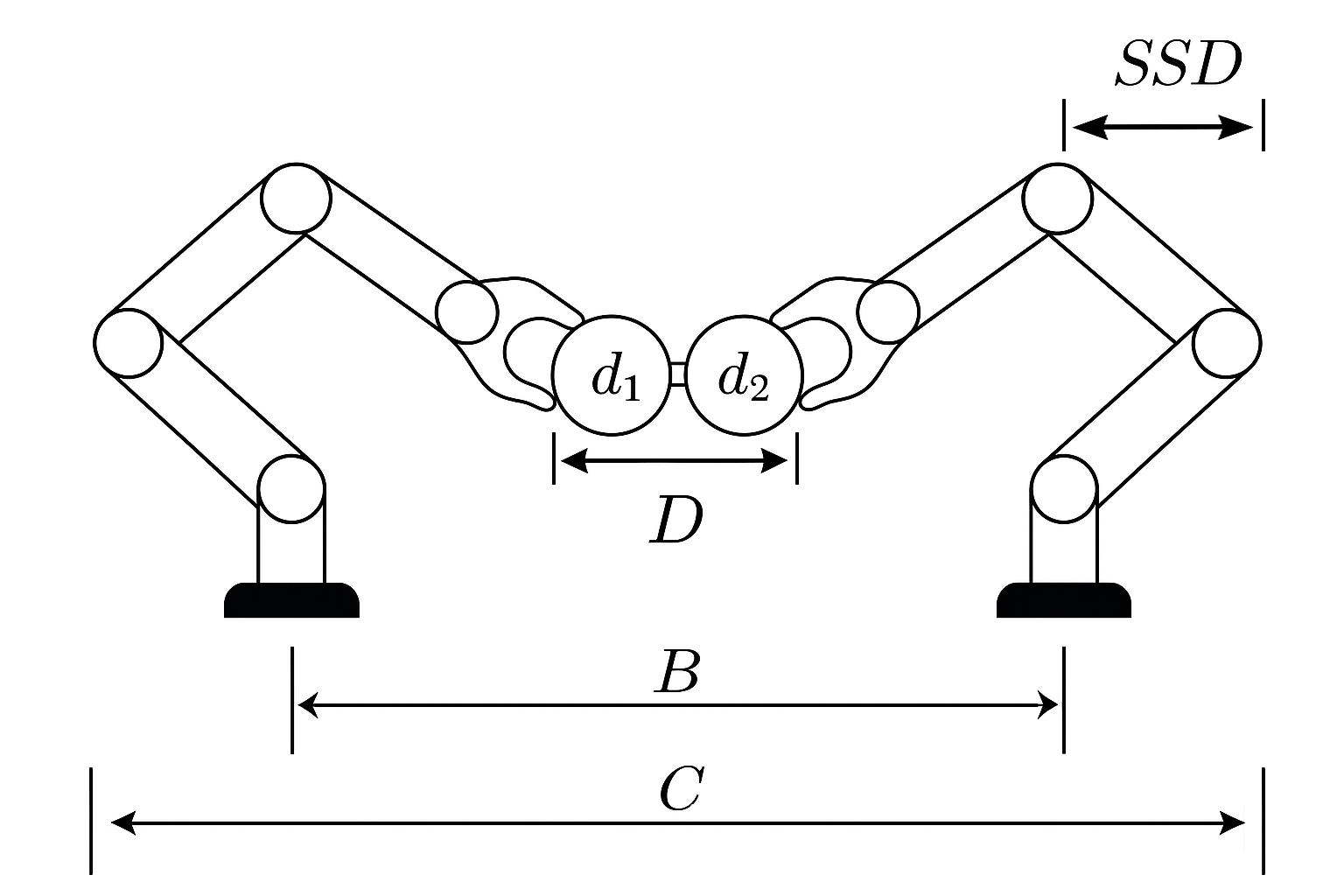}
\caption{Dimensions of a robotic cell. B is the distance between the bases of the robotic arms. C is the maximal width occupied by the robotic cell. SSD is the guaranteed maximal deviation of a single arm beyond its base. $d_1$ and $d_2$ are the maximal diameters of the sub-assemblies.
$D = d_1 + d_2$.}
\label{fig:robotic-cell-guarantee}
\end{figure}

As demonstrated in Section~\ref{sec:experiments}, valid trajectory placements using two robotic arms occur much more frequently than when using a single arm. Therefore, we can limit our analysis to trajectory placements on the plane that is normal to the line connecting the bases of the two arms and passes through the midpoint between them. We assume that after each step, the entire assembly is translated back to this plane (in practice, we combine this back-shift with the step itself). This allows the analysis of the robotic cell width to depend solely on the lengths of the robotic arms ($arm\_length$), the distance between the bases of the arms ($B$), and the combined diameters of the assembly pieces ($D$) (see Fig.~\ref{fig:robotic-cell-guarantee}). The length $D$ can be calculated simply as the sum $d_1 + d_2$ of the diameters of the two sub-assemblies, though a tighter bound may assess the diameter of the assembly as a whole at each step. Let $C$ represent the width of the robotic cell. Then:
\begin{equation}
    C=\frac{B}{2}+\frac{D}{2}+arm\_length\;.
\end{equation}
We assume that the distance between the bases of the robotic arms is larger than the assembly diameter, $B \ge D$, otherwise further restrictions on the trajectory may be required. Hence $C$, the maximal width occupied by the robotic cell, as a function of $B$, obtains its minimum when $B = D$:
\begin{equation}
    C_{min}=D+arm\_length.
\end{equation}
Further, in case the placements of the arms in the robotic cell are pre-defined and we are only interested in the single-side deviation ($SSD$) of the robotic arms out of the given robotic cell, we receive:
\begin{equation}
    SSD=\frac{D}{4}-\frac{B}{4}+\frac{arm\_length}{2}.
\end{equation}

\section{Experiments and Results}
\label{sec:experiments}
In this section we demonstrate the advantage of using \emph{CPW-IK} and \emph{GPW-IK} with two robotic arms over a single arm. 
A video compiling executions of all assemblies is available on our project page.

\subsection{Dataset}
We used the tight assembly dataset presented in~\cite{livnat2025fullcycle}, and compared the results of using two arms to those using a single arm. The four assemblies in that dataset were selected from a larger dataset~\cite{tian2022assemble}, keeping one challenging sample from each family. These are tight assemblies that require non-trivial combination of translation and rotation in order to be assembled, hence proved to be the hardest for the algorithms tested. In this work, 
as a proof of concept, we demonstrate the solutions, and measure the results, of four representative samples.

\subsection{Implementation Details}
We used the TR-RRT~\cite{livnat2024tight} motion planning algorithm for finding a valid trajectory for the free-flying assemblies.

We demonstrate our methods on a pair of \emph{Universal Robots UR5e} robotic arms, equipped with \emph{Robotiq 2F-85} grippers. 

The rest of the process is as follows. We begin with a free-flying trajectory $\bar{\gamma}$ and continue with running CPW-IK for a single arm, CPW-IK for two arms, or GPW-IK (for two arms) seeking valid placements. We continued with valid placements trials until $100$ valid placements were found for each combination of an assembly and an algorithm. We then design the parts in \emph{Blender}, add a small mark or cutout at the desired grasping point, which is defined manually, and print them using \emph{Creality Ender-3 S1} $3D$ printer with $0.2$ mm resolution. Finally, we grasp the parts using the end effectors and execute the valid trajectories in the real world as can be seen in Fig.~\ref{fig:az-zoom-in-out} and Fig.~\ref{fig:16505}.

\begin{table}[h]
\centering
\caption{Path quality. The average Hausdorff distance over 100 valid placements, presented as total radians traveled per time step, for each assembly. \emph{Error reduction}: percentage decrease in average Hausdorff distance when using two arms compared to a single arm. The \emph{Average} column reports an average only for the improvement row (percentage gains); averages of the raw per-assembly values are not shown because they are not directly meaningful.}
\label{table:new-hausdorff}
\renewcommand{\arraystretch}{1.2} 
\scalebox{1.0}{
\begin{tabular}{ |c || c c c c c |}
    \hline
    \textbf{Assembly} & az & abc & 16505 &  06397 & \emph{Average}\\
    \hline
    Single-arm & 0.035 & 0.041 & 0.028 & 0.121 & ---  \\
    Two-arms & 0.027 & 0.023 & 0.017 & 0.103 & --- \\
    \hline
    \textbf{Error reduction} & $\textbf{22.9\%}$ & $\textbf{43.9\%}$ & $\textbf{39.3\%}$ & $\textbf{14.9\%}$ & $\textbf{30.2\%}$ \\
    \hline
\end{tabular}
}
\end{table}

\subsection{Results}

The experiments demonstrate three main findings. 
First, the total assembly time (\emph{makespan}) is consistently lower when using two robotic arms than with a single arm. 
Fig.~\ref{fig:makespan} presents the complete set of results, including the improvement in the best solution found, the distribution of valid solutions, and the median performance. 
Relative to the \emph{average} solutions, the makespan is reduced by $15.9\%$ for CPW-IK and by $50.2\%$ for GPW-IK, while the improvement over the \emph{best} solutions is $14.4\%$ for CPW-IK and $39.5\%$ for GPW-IK. 
Note that these best-solution improvements do not reflect the full potential of GPW-IK, as discussed below and in Table~\ref{table:success-rate}, because on average it explored less than one third of the placements examined by the single-arm baseline.

\begin{table}[h]
\centering
\renewcommand{\arraystretch}{1.2} 
\caption{The table presents successful placements rate. \emph{Improvement} is the ratio between the two-arm success rate algorithms and the single-arm one. The \emph{Average} column reports an average only for the improvement rows (multiplicative gains); averages of the raw per-assembly values are not shown because they are not directly meaningful.} 

\scalebox{1.0}{
\begin{tabular}{ |c || c c c c c |}
    \hline
    \textbf{Assembly} & az & abc & 16505 &  06397 & \emph{Average}\\
    \hline
    Single-arm & 0.021 & 0.124 & 0.194 & 0.033 & ---  \\
    CPW-IK & 0.225 & 0.253 & 0.286 & 0.235 & --- \\
    \hline
    \textbf{Improvement} & $\textbf{x10.71}$ & $\textbf{x2.04}$ & $\textbf{x1.47}$ & $\textbf{x7.12}$ & $\textbf{x5.34}$ \\
    \hline
    GPW-IK & 0.156 & 0.128 & 0.245 & 0.150 & --- \\
    \hline
    \textbf{Improvement} & $\textbf{x7.43}$ & $\textbf{x1.03}$ & $\textbf{x1.26}$ & $\textbf{x4.55}$ & $\textbf{x3.57}$ \\
    \hline

\end{tabular}
}
\label{table:success-rate}
\end{table}

Second, the quality of the paths, in terms of one-sided Hausdorff distance, is improved, as can be seen in Table~\ref{table:new-hausdorff}. This means that the actual trajectory traveled by the assembly pieces when manipulated using two robotic arms, is closer to the desired free-flying trajectory given, than in the case of a single arm. 

Third, the trajectory placement algorithm performs much faster for CPW-IK and GPW-IK relative to the single-arm algorithm, as demonstrated in Table~\ref{table:success-rate} in terms of frequency of valid trajectories. That was for finding 100 valid placements. This implies that if we provide equal running times to all algorithms (rather than wait for 100 valid placements), the improvement in assembly time is expected to be even larger (as the two-arm algorithms will have time to test a larger amount of placements).

\section{Discussion}

Our experiments confirm that coordinated dual-arm planning dramatically reduces assembly makespan and improves trajectory accuracy compared to a single-arm baseline. 

This work introduced GPW-IK, a new dual-arm planning algorithm with a provable $>50\%$ reduction in average assembly time and demonstrated its effectiveness on physical hardware.

These results highlight the practical relevance of our framework for Industry~4.0 manufacturing cells and motivate its adoption in time-critical assembly lines.

\begin{figure}[H]
\centering
\includegraphics[width=0.5\linewidth]{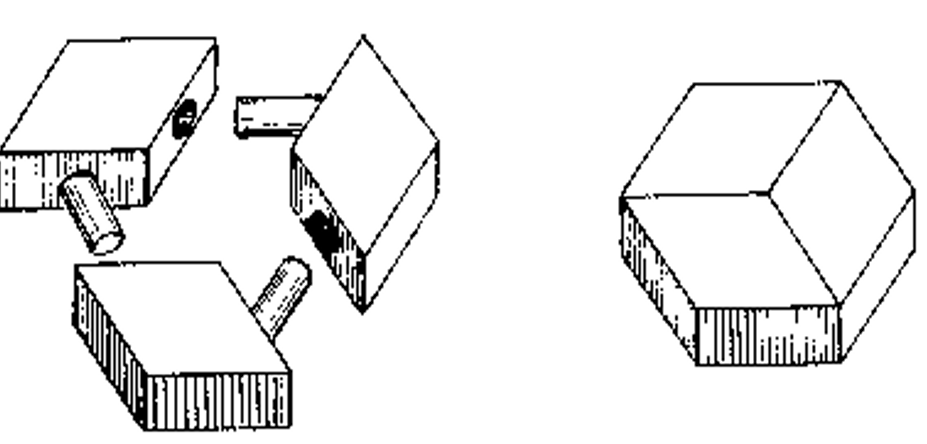}
\caption{Conceptual illustration of a three-handed assembly (adapted from johnrausch.com); future work will extend our method to such cases.}
\label{fig:3-handed}
\end{figure}

We continue with this research in two directions. First, is to further accelerate the simultaneous work of two arms by leveraging global information rather than only that of the current segment. Second, there are assembly operations involving three objects, which require coordinated manipulation of two arms in order to complete the assembly (see Fig.~\ref{fig:3-handed}). We plan to use the algorithm and software package developed for this project to solve such three-handed assembly problems.

\bibliographystyle{IEEEtran}
\bibliography{bibliography}

\end{document}